\documentclass[sigconf]{acmart}
\renewcommand\footnotetextcopyrightpermission[1]{}  
\AtBeginDocument{%
  }

\usepackage{balance}
\usepackage{xcolor}

\thanks{
This is the author’s accepted manuscript of a paper accepted to the
**Causality in Software Engineering (CauSE) 2025 Workshop**, held in conjunction with the
**33rd ACM Joint European Software Engineering Conference and Symposium on the Foundations of Software Engineering (ESEC/FSE 2025)**,
June 23–28, 2025, Trondheim, Norway.
The final version will appear in the ACM Digital Library: \url{https://doi.org/10.1145/3696630.3731615}
}

\begin{document}

\title{Causality-Driven Neural Network Repair:\\ Challenges and Opportunities}


\author{Fatemeh Vares}
\affiliation{%
  \institution{George Mason University}
  \city{Fairfax}
  \state{Virginia}
  \country{USA}
}
\email{fvares@gmu.edu}

\author{Brittany Johnson}
\affiliation{%
  \institution{George Mason University}
  \city{Fairfax}
  \state{Virginia}
  \country{USA}
}
\email{johnsonb@gmu.edu}

\newcommand{\BJohnson}[1]{\textcolor{purple}{{\bfseries [[#1]]}}}
\newcommand{\FVares}[1]{\textcolor{red}{{\bfseries [[#1]]}}}


\begin{abstract}
Deep Neural Networks (DNNs) often rely on statistical correlations rather than causal reasoning, limiting their robustness and interpretability. 
While testing methods can identify failures, effective debugging and repair remain challenging. 
This paper explores causal inference as an approach primarily for DNN repair, leveraging causal debugging,
counterfactual analysis, and structural causal models (SCMs) to identify and correct failures. 
We discuss in what ways these techniques support fairness, adversarial robustness, and backdoor mitigation by providing targeted interventions. 
Finally, we discuss key challenges, including scalability, generalization, and computational efficiency, and outline future directions for integrating causality-driven interventions to enhance DNN reliability.
\end{abstract}

\ccsdesc[500]{Software and its engineering~Software notations and tools}
\ccsdesc[500]{Computing methodologies~Artificial intelligence}

\keywords{Causal Inference, Neural Network Repair, Deep Neural Networks (DNNs), Causal Debugging, Explainability in Deep Learning}


\maketitle
\settopmatter{printfolios=false}

\section{Introduction}
 The rapid and widespread adoption of Deep Neural Networks (DNNs) has raised concerns about their reliability and robustness. 
 Various testing approaches have successfully identified misbehavior in DNNs \cite{sohn2023arachne}, but methods for correcting these errors remain unclear. 
In traditional software programs, debugging and repair rely on well-defined notions of causality, such as control and data dependencies \cite{johnson2020causal, ibrahim2020actual}.
 However, neural networks differ significantly in this regard. 
 Incorrect predictions in DNNs do not stem from a single erroneous component but rather from a combination of factors, including the structure of the network, the properties of input data, and the complex interactions among neurons. 
 As a result, attributing responsibility to specific elements and applying targeted modifications is highly challenging \cite{sun2022causality}.

Modern deep learning models heavily rely on statistical correlations rather than genuine causal relationships. 
This limitation hinders their generalization, especially in domains that require high adaptability, such as medicine and autonomous driving \cite{li2020deep, han2018robust}.
Existing approaches, including self-supervised learning \cite{liu2021generative}, semi-supervised learning \cite{zhang2021analysis}, and reinforcement learning \cite{wang2022deep}, primarily focus on optimizing model performance based on large amounts of labeled data or extensive interactions with the environment. 
However, these methods do not explicitly incorporate causal reasoning, which is capable of improving model robustness and interpretability~\cite{jiao2024causal}.

Causal learning has emerged as a promising direction to address these challenges by distinguishing true causal relationships from spurious correlations \cite{cheng2021causal}. This field consists of \textit{causal discovery}, which identifies causal structures within data~\cite{spirtes2016causal}, and \textit{causal inference}, which quantifies the strength of causal effects assuming an existing causal structure~\cite{spirtes2001causation}. 
While causal discovery methods require extensive data and computational resources, causal inference allows for a more efficient estimation of causal effects, enhancing model adaptability to new environments.

Recent advancements in causal inference have demonstrated its potential in deep learning applications, including visual representation learning \cite{liu2022causal}, video processing \cite{liu2022causal, zhang2020shadow}, vision-language modeling \cite{buch2022revisiting}, interpretability of deep models \cite{su2022explainable, tjoa2020survey}, and natural language processing (NLP)~\cite{wood2018challenges}. 
Cai et al. \cite{cai2024and} proposed CADE, a causality-driven adversarial attack framework for DNNs, particularly CNNs like ResNet-50 and VGG-16. It was tested on Pendulum, CelebA, and SynMeasurement datasets. Unlike traditional attacks, CADE used Structural Causal Models (SCMs) to identify where and how to attack, ensuring more realistic adversarial examples. It followed Pearl’s \cite{pearl2009causal} counterfactual framework (abduction, action, prediction) to model interventions and generate counterfactual adversarial samples.

Zhang et al. \cite{zhang2021causaladv} further extended this concept with CausalAdv, a causal-inspired adversarial distribution alignment method to enhance the robustness of deep neural networks (DNNs) against adversarial attacks. By constructing a causal graph, the authors identified spurious correlations as the key factor behind adversarial vulnerability. CausalAdv aimed to mitigate this by aligning the adversarial and natural distributions, reducing the model's reliance on non-causal features. The study focused on CNN architectures like ResNet-18 and WRN-34-10, evaluating performance on MNIST, CIFAR-10, and CIFAR-100 datasets against adversarial attacks such as FGSM, PGD, C\&W, and AutoAttack.

These studies suggest that incorporating causality into deep learning debugging and repair can lead to more reliable and generalizable models. 
This paper explores the role of causal inference in deep learning and discusses its implications for improving model robustness in dynamic environments. The rest of this paper is structured as follows: Section 2 surveys feature-level and neuron-level causal repair techniques. Section 3 discusses associated challenges and research opportunities. Section 4 concludes the paper with future directions.


\begin{table*}[ht]
  \caption{Comparison of Causality-Based Neural Network Repair Approaches}
  \label{tab:repair_comparison}
  \centering
  \begin{tabular}{l l p{10cm}} 
    \toprule
    \textbf{Method} & \textbf{Approach} & \textbf{Limitations} \\
    \midrule
    \cite{mcqueary2024py} & Counterfactual debugging & Use of assistive sample generation instead of data augmentation; evaluation on a limited dataset (MNIST) \\
    \cite{sun2022causality} & Structural causal models (SCMs) & a trade-off between fixing misbehavior and maintaining the model’s original accuracy, computationally expensive optimization process using PSO algorithm\\
    \cite{liu2024towards} & Counterfactual tracing & Computational complexity; reduced accuracy in some cases; applicability to larger models and real-world architectures remains uncertain \\
    \cite{zhang2021causaladv} & Causal graph-based learning & Limited large-scale evaluation\\
    \cite{pawlowski2020deep} & Deep structural causal modeling & Limited by assumptions of complete observability, challenges in training, difficulty in counterfactual validation\\ 
    \cite{narendra2018explaining} & Causal explanation for CNNs & Limited expressiveness of structural equations, Limited transformation methods, fixed dataset assumption \\ 
    \cite{scholkopf2021toward} & Causal representation learning & Scalability issues; difficulty learning causal variables; limitations in causal reinforcement learning  \\ 
    \cite{berrevoets2024causal} & Causal deep learning framework & Partial causal knowledge dependence, parametric complexity, confounding in temporal models, and uncertainty in causal structure assumptions \\ 
    \bottomrule
  \end{tabular}
\end{table*}

\section{Causal Inference for Neural Network Repair}
Conventional DNN repair methods suffer from several limitations, including reliance on statistical correlations \cite{zhang2021causaladv, tang2020long}, poor generalization across domains \cite{cai2024and}, and lack of interpretability\cite{sun2022causality}. To address these issues, researchers have explored causal inference techniques for diagnosing and repairing neural network failures.
Neural network repair aims to improve the robustness and reliability of deep learning models by addressing identified failure points. However, conventional repair techniques, such as adversarial training, fine-tuning, and weight pruning, primarily rely on statistical correlations rather than causal mechanisms, often resulting in non-generalizable fixes. Causal inference introduces a structured methodology to diagnose and repair DNN failures by distinguishing spurious correlations from genuine causal dependencies \cite{jiao2024causal}.
To address these challenges, we categorize causal inference-based repair methods into two main approaches: feature-level interventions, which focus on modifying input-output dependencies, and neuron-level interventions, which involve adjustments to the internal structure of the model.

\subsection{Feature-Level Causal Interventions}
Feature-level interventions focus on adjusting input-output dependencies by eliminating spurious correlations in neural networks. Tang et al. \cite{tang2020long} proposed a causal inference framework to address long-tailed classification by mitigating momentum bias in SGD. Using de-confounded training and total direct effect (TDE) inference, it removed spurious correlations while preserving beneficial feature relationships. Py-Holmes \cite{mcqueary2024py} facilitated debugging in DNNs by generating counterfactual inputs to diagnose failures. The system perturbed input tensors using loss gradients and filtered assistive samples based on neuron activation similarity, pinpointing critical model failures.
SCM-based causal repair \cite{sun2022causality} applied Structural Causal Models (SCMs) and Average Causal Effect (ACE) analysis to eliminate biases in input data relationships. This method enhanced fairness, mitigated backdoor attacks, and improved robustness in structured datasets. CausalAdv \cite{zhang2021causaladv} introduced causal graph-based learning to mitigate adversarial vulnerabilities by aligning adversarial and natural distributions, reducing the model’s reliance on non-causal features. However, CausalAdv faced challenges in large-scale evaluation, as causal graphs could be computationally expensive to construct and optimize.
Deep Structural Causal Modeling (DSCM) \cite{pawlowski2020deep} leveraged causal representations to improve model interpretability and fairness but was limited by assumptions of complete observability, challenges in training, and difficulty in counterfactual validation. Causal representation learning \cite{scholkopf2021toward} aimed to enhance generalization by identifying causal factors in neural networks. However, scalability issues, difficulty in learning causal variables, and limitations in causal reinforcement learning remained significant challenges. Although quantitative comparison is outside the scope of this short paper, Table 1 qualitatively contrasts representative methods based on their main limitations, highlighting where future evaluation efforts are needed.

\subsection{Neuron-Level Causal Interventions}
Neuron-level interventions targeted specific components of deep networks by detecting and modifying faulty neurons to improve model performance and reliability. CARE (CAusality-based REpair) \cite{sun2022causality} identified and corrected faulty neurons using SCMs and ACE estimation. It employed Particle Swarm Optimization (PSO) to optimize neuron weights while preserving accuracy across fairness, security, and backdoor removal tasks. CCBR (Counterfactual Causality-Based Repair) \cite{liu2024towards} applied counterfactual tracing and NSGA-III optimization to detect and adjust faulty neurons. The framework modeled the neural network as a Counterfactual Structural Causal Model (CSCM), improving security properties and fairness while reducing vulnerabilities.
Causal explanations for CNNs \cite{narendra2018explaining} analyzed the causal influence of individual filters in convolutional networks, offering greater transparency in model decision-making. However, these methods were constrained by limited expressiveness of structural equations, reliance on transformation methods, and fixed dataset assumptions. The causal deep learning framework \cite{berrevoets2024causal} integrated causality into deep learning but faced challenges related to partial causal knowledge dependence, parametric complexity, confounding in temporal models, and uncertainty in causal structure assumptions.
\\ Despite these advancements, several key challenges remain in effectively implementing causality-driven neural network repair. Addressing these challenges is crucial for making these methods practical and scalable in real-world deep learning applications, as discussed in the next section.
\section{Challenges \& Opportunities}
Causality-driven neural network repair presents several significant challenges that must be addressed to enable broader adoption in deep learning. However, these challenges also open up opportunities for innovation and improvement. Below, we categorize key challenges and discuss the corresponding opportunities to advance this field.
\subsection{Computational Scalability}
Structural Causal Models (SCMs) and counterfactual reasoning techniques often require extensive computational resources \cite{sun2022causality}. As the number of variables and dependencies grows, the complexity of these models increases exponentially, making them impractical for large-scale neural networks. 
Advances in hardware acceleration, such as optimized GPU and TPU implementations, can help mitigate computational costs. Research by Zhang et al. \cite{zhang2021causaladv} suggests that alternative approaches like causal adversarial robustness frameworks, which integrate causal learning with adversarial defenses, can achieve computational efficiency without significantly increasing complexity. Furthermore, pruning and compression strategies, as explored by Liu et al. \cite{liu2024towards}, can enable scalable causal inference without compromising accuracy. Developing lightweight causal models that balance expressiveness and computational efficiency remains a promising research direction.
\subsection{Causal Discovery in High-Dimensional Data}
Extracting meaningful causal relationships from complex, multi-dimensional datasets is inherently difficult. Traditional causal inference methods struggle to scale efficiently in environments with high feature interdependencies, leading to unreliable causal conclusions. 
Leveraging deep learning itself for causal discovery through representation learning can improve scalability. Hybrid models that integrate data-driven learning with domain knowledge, as explored by Berrevoets et al. \cite{berrevoets2024causal}, could enhance causal structure identification. Additionally, self-supervised learning methods, such as those investigated by Tang et al. \cite{tang2020long}, can help uncover causal patterns without the need for extensive labeled datasets, making causal discovery more feasible in high-dimensional settings.
\subsection{Optimization Trade-offs}
Causal interventions must balance multiple objectives, including robustness, interpretability, and predictive accuracy. While some methods improve robustness against adversarial attacks, they may inadvertently reduce model accuracy. Sun et al. \cite{sun2022causality} highlight that while causal repair methods enhance security and fairness, they often lead to trade-offs that must be carefully managed to prevent degradation in overall performance.
Multi-objective optimization frameworks, such as those proposed by Liu et al. \cite{liu2024towards}, can help navigate these trade-offs by balancing causal corrections with model performance. Reinforcement learning-based adaptive interventions can fine-tune causal adjustments dynamically, optimizing both fairness and accuracy without significant performance degradation. Zhang et al. \cite{zhang2021causaladv} also emphasize that causal-inspired adversarial methods can mitigate the robustness-accuracy trade-off by aligning adversarial and natural data distributions effectively.
\subsection{Lack of Standardized Benchmarks}
Unlike adversarial robustness, which has widely accepted evaluation metrics, causal repair lacks a universally recognized framework for assessing its effectiveness. This makes it difficult to compare different approaches and measure improvements consistently. Spirtes et al.\cite{spirtes2001causation} argue that the absence of standardized benchmarks limits progress in causal inference applications, as researchers struggle to validate their methodologies against common criteria.
Establishing standardized datasets and evaluation metrics specific to causal repair can drive progress in the field. Collaborative benchmarking efforts across research institutions and industry, as suggested by \cite{pearl2009causal}, can help define reliable assessment methods, ensuring fair comparisons of different causal repair techniques. Additionally, incorporating causal evaluation metrics within existing deep learning frameworks can streamline their adoption in practical applications.
\subsection{Integration with Deep Learning Architectures}
Most modern deep learning frameworks are designed for correlation-based learning rather than causal reasoning. Integrating causal inference into existing architectures requires novel methodologies that seamlessly integrate with current optimization techniques and training pipelines. Sun et al. \cite{sun2022causality} discuss that current neural network repair frameworks require extensive modifications to standard architectures, limiting their widespread adoption.
Developing modular causality-aware layers that can be plugged into standard neural network architectures would facilitate smoother integration. Advances in differentiable causal inference techniques, as explored by Berrevoets et al. \cite{berrevoets2024causal}, can bridge the gap between deep learning and causal reasoning, allowing models to incorporate causal understanding naturally. Future research should explore hybrid neural architectures that inherently support causal inference, reducing the reliance on post-hoc causal adjustments.
\\ By addressing these challenges, causality-driven neural network repair can become more scalable, interpretable, and effective. Future research should focus on refining these methodologies to enable practical and impactful applications in deep learning.

\section{Conclusion}
Causal inference provides a structured, interpretable, and targeted approach for DNN repair. Conventional repair methods suffer from a reliance on statistical correlations, leading to fragile fixes. By incorporating causal debugging, counterfactual analysis, and SCM-based interventions, researchers have demonstrated improvements in robustness against adversarial attacks, mitigation of spurious correlations, and enhanced generalization across domains.
Future research should prioritize scalable causal discovery techniques, integration of causal reasoning into mainstream deep learning architectures, and the development of standardized benchmarks for evaluating causal repair methods \cite{jiao2024causal}. Additionally, the combination of causal repair with reinforcement learning and self-supervised learning could further enhance adaptability in dynamic environments. Given the increasing deployment of deep learning models in safety-critical domains, causality-driven interventions will be essential for ensuring long-term reliability and trustworthiness in AI systems.

%
\bibliographystyle{ACM-Reference-Format} 

\end{document}